\documentclass[10pt,twocolumn,letterpaper]{article}

\usepackage{iccv}
\usepackage{times}
\usepackage{epsfig}
\usepackage{graphicx}
\usepackage{amsmath}
\usepackage{amssymb}
\usepackage{subfig}

\usepackage[pagebackref=true,breaklinks=true,letterpaper=true,colorlinks,bookmarks=false]{hyperref}

\iccvfinalcopy 

\def\iccvPaperID{81} 

\ificcvfinal\fi
\begin{document}

\setlength{\textfloatsep}{8pt}
\setlength{\dbltextfloatsep}{8pt}
\setlength{\abovedisplayskip}{4pt}
\setlength{\belowdisplayskip}{4pt}

\title{DirectDrive: Direct Perception for Self-driving using Deep Learning}
\title{DeepDrive: An Intermediate Compact Driving Representation for Vision-based Autonomous Driving}
\title{DeepDrive: An Intermediate Representation for Autonomous Driving}
\title{Learning an Intermediate Representation for Autonomous Driving}
\title{Learning an Intermediate Representation for Autonomous Driving}
\title{Learning Direct Perception for Autonomous Driving using Deep Learning}
\title{Learning Direct Perception Representation for Autonomous Driving}
\title{Direct Perception for Autonomous Driving: Learning Affordance Representation}
\title{Learning Affordance for Direct Perception in Autonomous Driving}
\title{DeepDriving: Learning Affordance for Direct Perception in Autonomous Driving}

\author{
Chenyi Chen
\quad
Ari Seff
\quad
Alain Kornhauser
\quad
Jianxiong Xiao\\
Princeton University\\
\href{http://deepdriving.cs.princeton.edu}{http://deepdriving.cs.princeton.edu}
}

\maketitle

\begin{abstract}
Today, there are two major paradigms for vision-based autonomous driving systems:
mediated perception approaches that parse an entire scene to make a driving decision,
and behavior reflex approaches that directly map an input image to a driving action by a regressor.
In this paper, we propose a third paradigm:
a direct perception approach to estimate the affordance for driving.
We propose to map an input image to a small number of key perception indicators
that directly relate to the affordance of a road/traffic state for driving. 
Our representation provides a set of compact yet complete descriptions of the scene to enable a simple controller to drive autonomously.
Falling in between the two extremes of mediated perception and behavior reflex,
we argue that our direct perception representation provides the right level of abstraction.
To demonstrate this,
we train a deep Convolutional Neural Network using recording from 12 hours of human driving in a video game
and show that our model can work well
to drive a car in a very diverse set of virtual environments.
We also train a model for car distance estimation on the KITTI dataset.
Results show that our direct perception approach can generalize well to real driving images.
Source code and data are available on our project website.
\end{abstract}

\vspace{-2mm}
\section{Introduction}

In the past decade, significant progress has been made in autonomous driving.
To date, most of these systems can be categorized into two major paradigms:
mediated perception approaches and behavior reflex approaches.

\begin{figure}[t]
  \centering
    \includegraphics[width=1\linewidth]{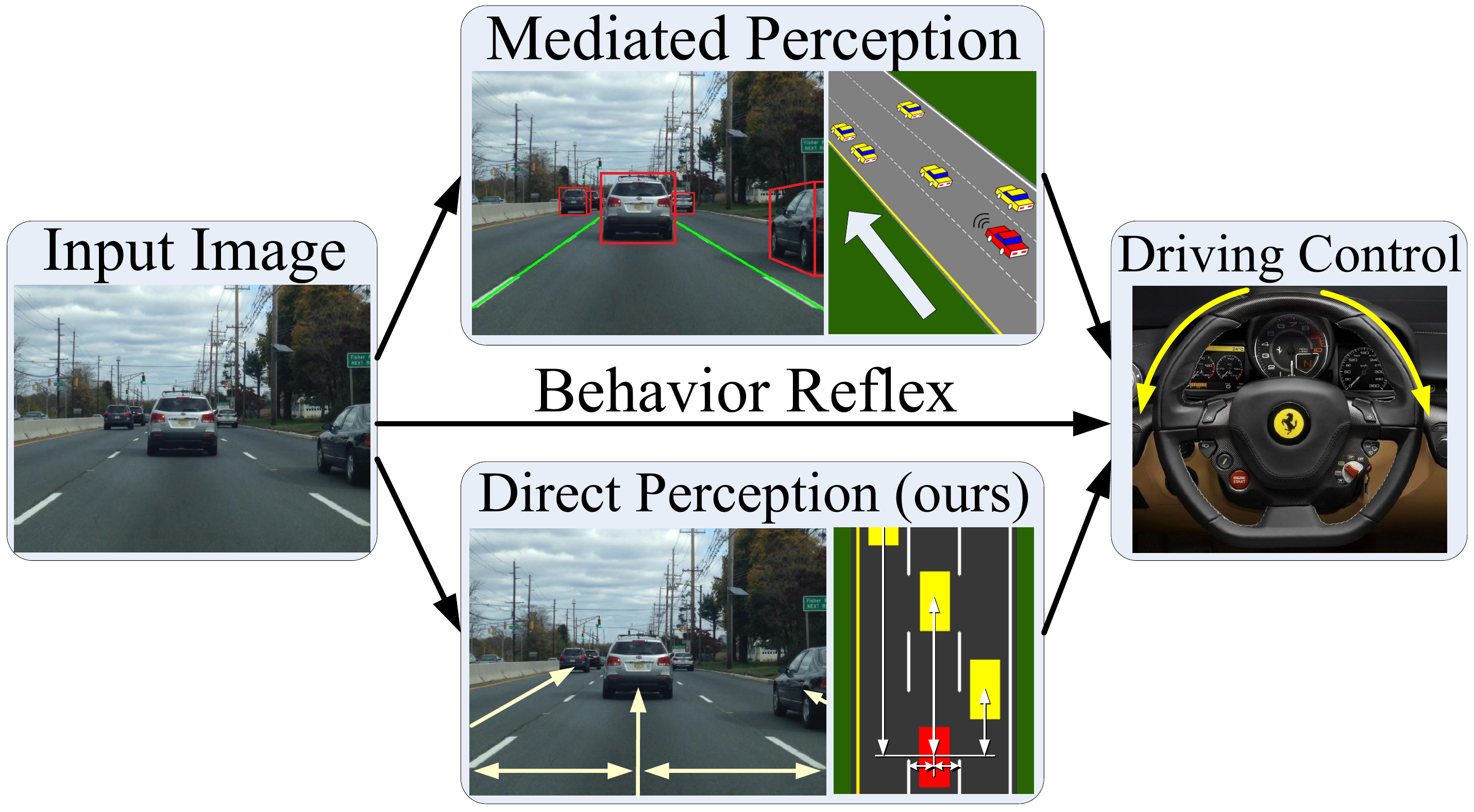}

\vspace{-2mm}
    \caption{\bf Three paradigms for autonomous driving.}
    \label{approaches}






\end{figure}

\begin{figure*}[t]
\vspace{-4mm}

    \subfloat[one-lane]{\label{fig2a}\includegraphics[width=0.16\linewidth]{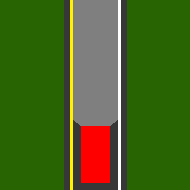}}~
    \subfloat[two-lane, left]{\label{fig2b}\includegraphics[width=0.16\linewidth]{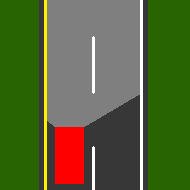}}~
    \subfloat[two-lane, right]{\label{fig2c}\includegraphics[width=0.16\linewidth]{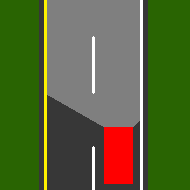}}~
    \subfloat[three-lane]{\label{fig2e}\includegraphics[width=0.16\linewidth]{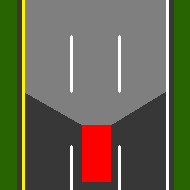}}~
    \subfloat[inner lane mark.]{\label{fig3a}\includegraphics[width=0.16\linewidth]{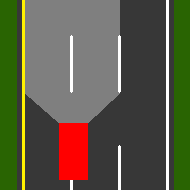}}~
    \subfloat[boundary lane mark.]{\label{fig3b}\includegraphics[width=0.16\linewidth]{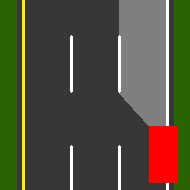}}

    \vspace{-3mm}
    \caption{{\bf Six examples of driving scenarios from an ego-centric perspective.} The lanes monitored for making driving decisions are marked with light gray.  }
    \label{fig2}
\end{figure*}

\vspace{2mm}{\bf Mediated perception approaches} \cite{ullman1980against} involve multiple sub-components for recognizing driving-relevant objects, such as lanes, traffic signs, traffic lights, cars, pedestrians, etc. \cite{geiger2013vision}. The recognition results are then combined into a consistent world representation of the car's immediate surroundings (Figure \ref{approaches}). To control the car, an AI-based engine will take all of this information into account before making each decision.
Since only a small portion of the detected objects are indeed relevant to driving decisions, this level of total scene understanding may add unnecessary complexity to an already difficult task. Unlike other robotic tasks, driving a car only requires manipulating the direction and the speed. This final output space resides in a very low dimension, while mediated perception computes a high-dimensional world representation, possibly including redundant information. Instead of detecting a bounding box of a car and then using the bounding box to estimate the distance to the car, why not simply predict the distance to a car directly? After all, the individual sub-tasks involved in mediated perception are themselves considered open research questions in computer vision. 
Although mediated perception encompasses the current state-of-the-art approaches for autonomous driving, 
most of these systems 
have to rely on laser range finders, GPS, radar and very accurate maps of the environment
to reliably parse objects in a scene.
Requiring solutions to many open challenges for general scene understanding in order to solve the simpler car-controlling problem 
unnecessarily increases the complexity and the cost of a system.

{\bf Behavior reflex approaches} 
construct a direct mapping from the sensory input to a driving action.
This idea dates back to the late 1980s when \cite{pomerleau1989alvinn,pomerleau1992neural}
used a neural network to construct a direct mapping from an image to steering angles.
To learn the model, a human drives the car along the road while the system records the images and steering angles as the training data.
Although this idea is very elegant,
it can struggle to deal with traffic and complicated driving maneuvers for several reasons.
Firstly,
with other cars on the road, even when the input images are similar,
different human drivers may make completely different decisions,
which results in an ill-posed problem that is confusing when training a regressor.
For example, with a car directly ahead, one may choose to follow the car, to pass the car from the left, or to pass the car from the right.
When all these scenarios exist in the training data,
a machine learning model will have difficulty deciding what to do given almost the same images.
Secondly,
the decision-making for behavior reflex is too low-level.
The direct mapping cannot see a bigger picture of the situation.
For example, from the model's perspective, passing a car and switching back to a lane are just a sequence of very low level decisions for turning the steering wheel slightly
in one direction and then in the other direction for some period of time.
This level of abstraction fails to capture what is really going on, and it increases the difficulty of the task unnecessarily.
Finally, because the input to the model is the whole image,
the learning algorithm must determine which parts of the image are relevant.
However, the level of supervision to train a behavior reflex model, i.e. the steering angle,
may be too weak to force the algorithm to learn this critical information.

We desire a representation that directly
predicts the affordance for driving actions,
instead of visually parsing the entire scene or blindly mapping an image to steering angles.
In this paper, we propose a {\bf direct perception approach} \cite{gibson2013ecological} for autonomous driving
-- a third paradigm that falls in between mediated perception and behavior reflex.
We propose to learn a mapping from an image to several meaningful affordance indicators of the road situation,
including the angle of the car relative to the road, the distance to the lane markings, and the distance to cars in the current and adjacent lanes.
With this compact but meaningful affordance representation as perception output,
we demonstrate that a very simple controller can then make driving decisions at a high level and drive the car smoothly.

Our model is built upon the state-of-the-art deep Convolutional Neural Network (ConvNet) framework
to automatically learn image features for estimating affordance related to autonomous driving.
To build our training set, we ask a human driver
to play a car racing video game TORCS
for 12 hours while recording the screenshots and the corresponding labels.
Together with the simple controller that we design, our model can make meaningful predictions for affordance indicators
and autonomously drive a car in different tracks of the video game,
under different traffic conditions and lane configurations.
At the same time, it enjoys a much simpler structure than the typical mediated perception approach.
Testing our system on car-mounted smartphone videos and the KITTI dataset \cite{geiger2013vision} demonstrates good real-world perception as well.
Our direct perception approach provides a
compact, task-specific affordance description for scene understanding in autonomous driving.


\begin{figure*}[t]

\vspace{-4mm}

\hspace{-1mm}%
\subfloat[angle]{\label{fig3c}\includegraphics[height=0.22\linewidth]{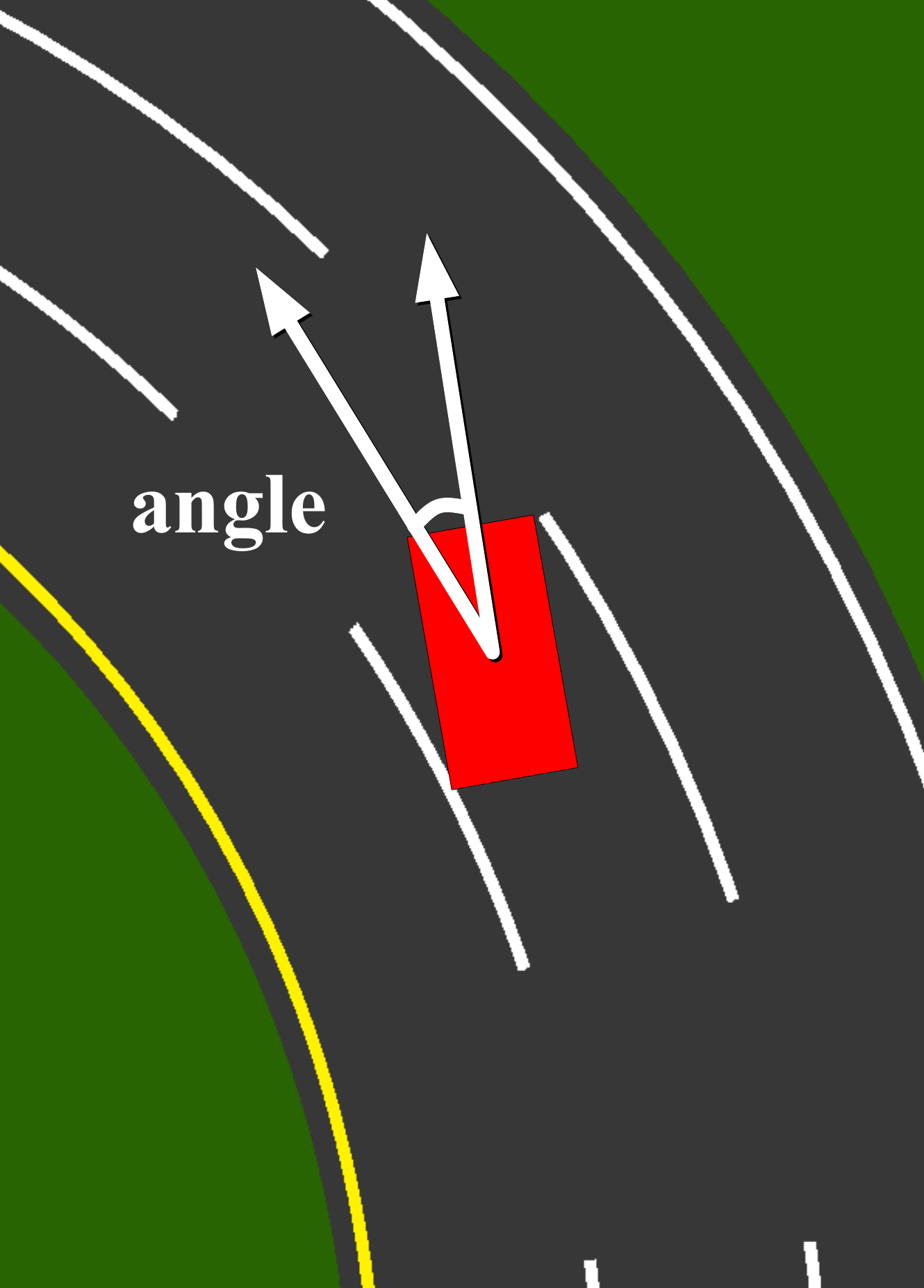}}~~%
\subfloat[in lane: toMarking]{\label{fig4b}\includegraphics[height=0.22\linewidth]{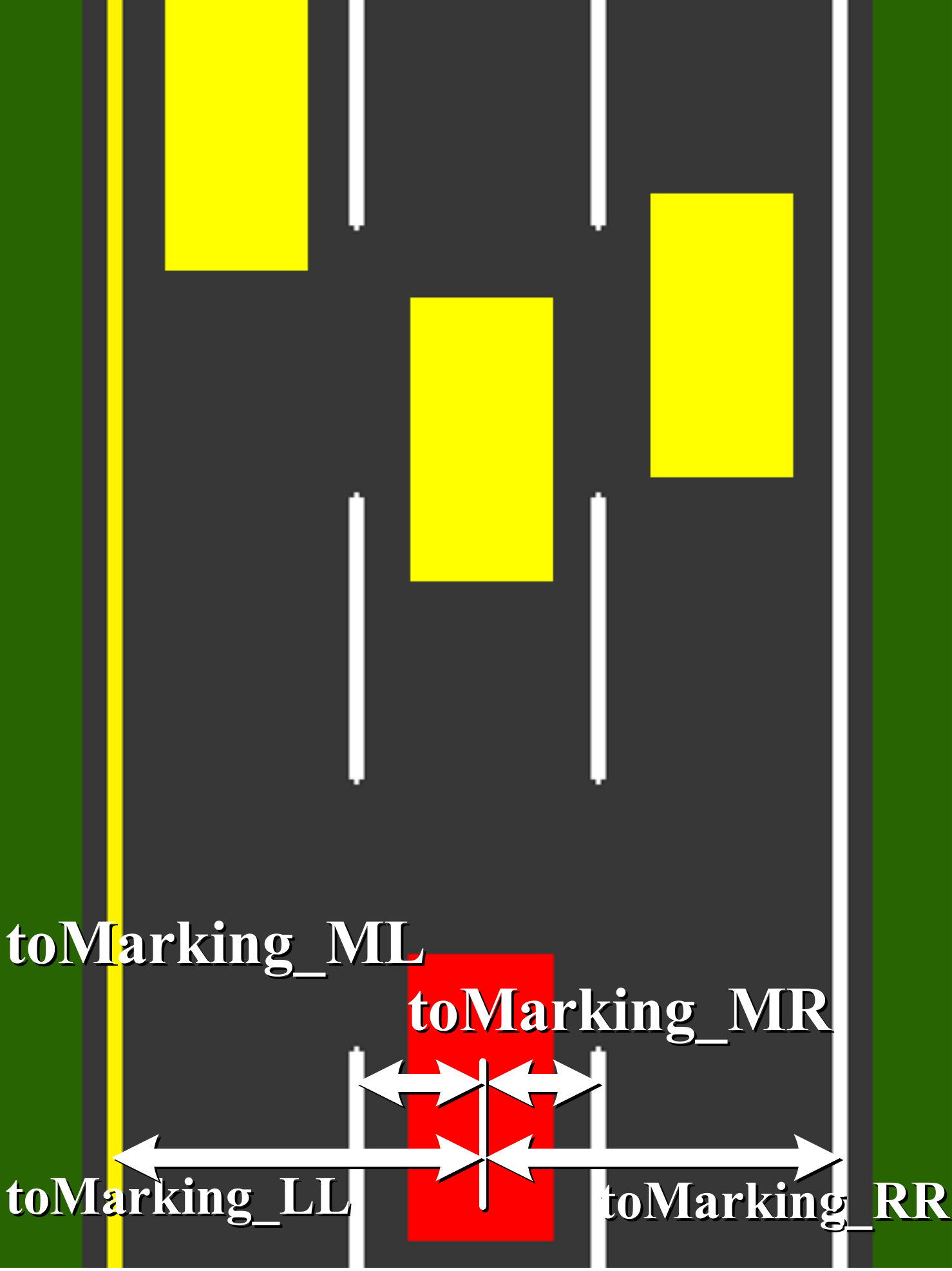}}~~%
\subfloat[in lane: dist]{\label{fig4c}\includegraphics[height=0.22\linewidth]{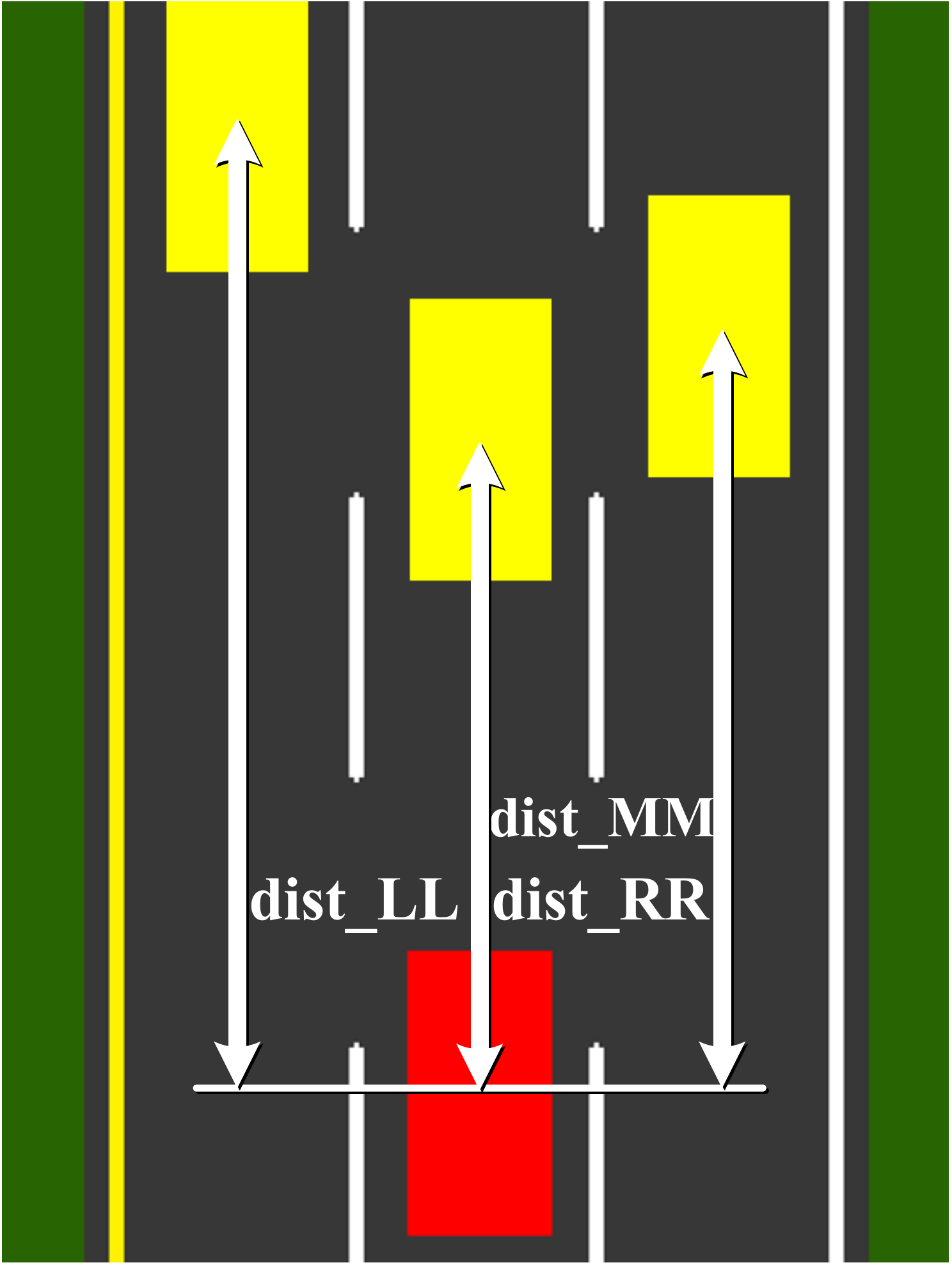}}~~%
\subfloat[on mark.: toMarking\hspace{-5mm}]{\label{fig4d}\includegraphics[height=0.22\linewidth]{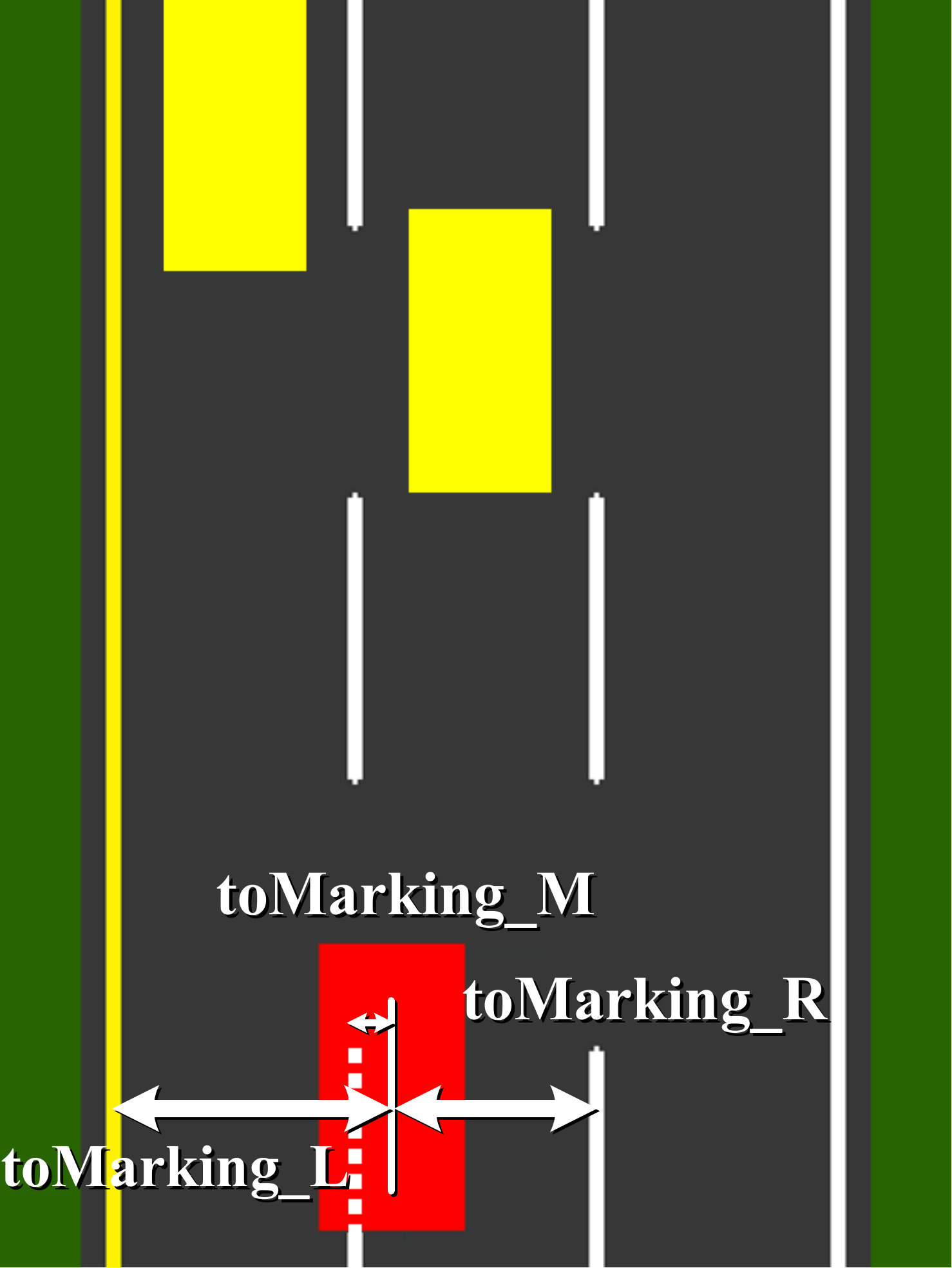}}~~%
\subfloat[on marking: dist]{\label{fig4e}\includegraphics[height=0.22\linewidth]{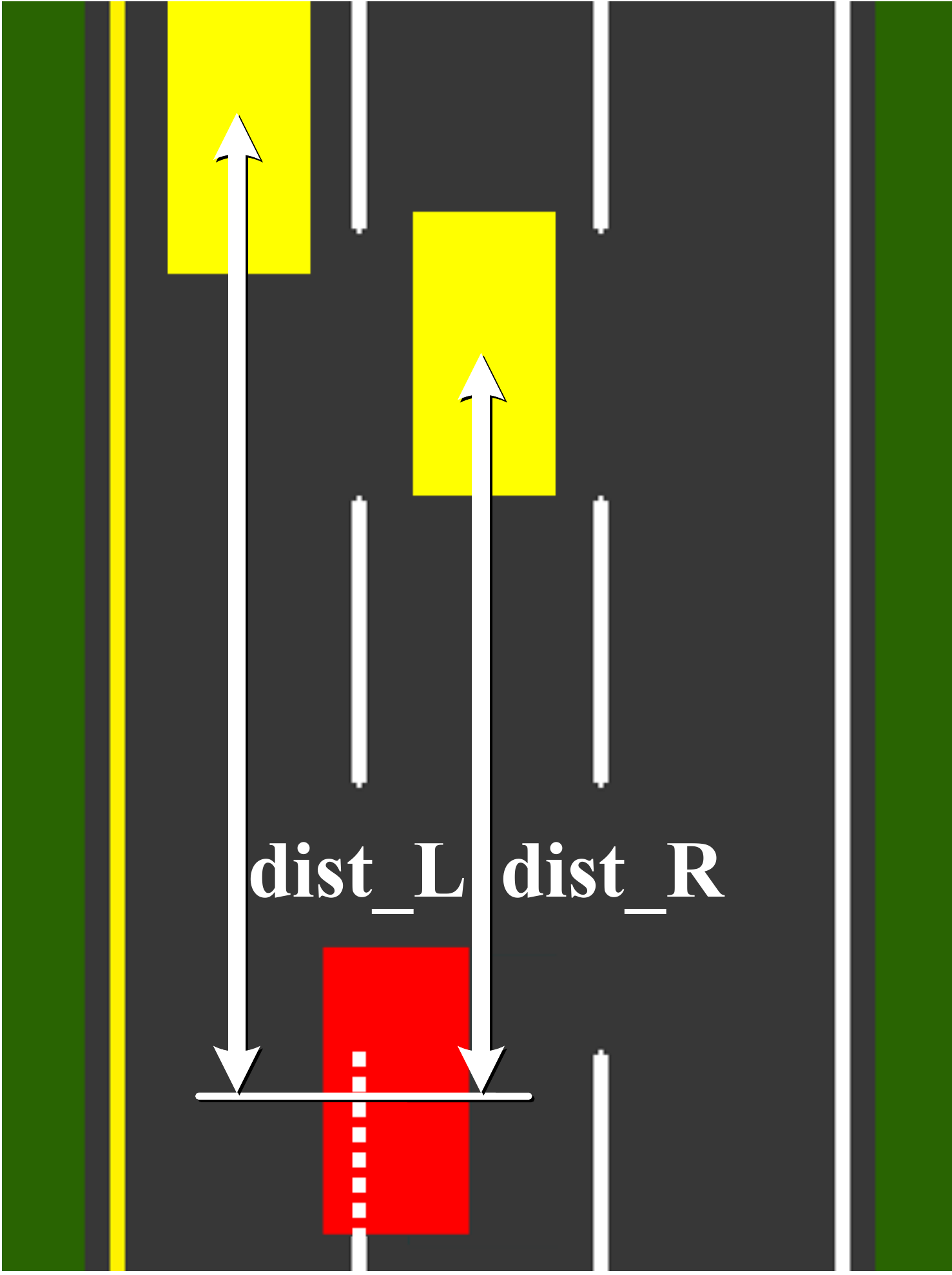}}~~%
\subfloat[overlapping area]{\label{fig5}\includegraphics[height=0.22\linewidth]{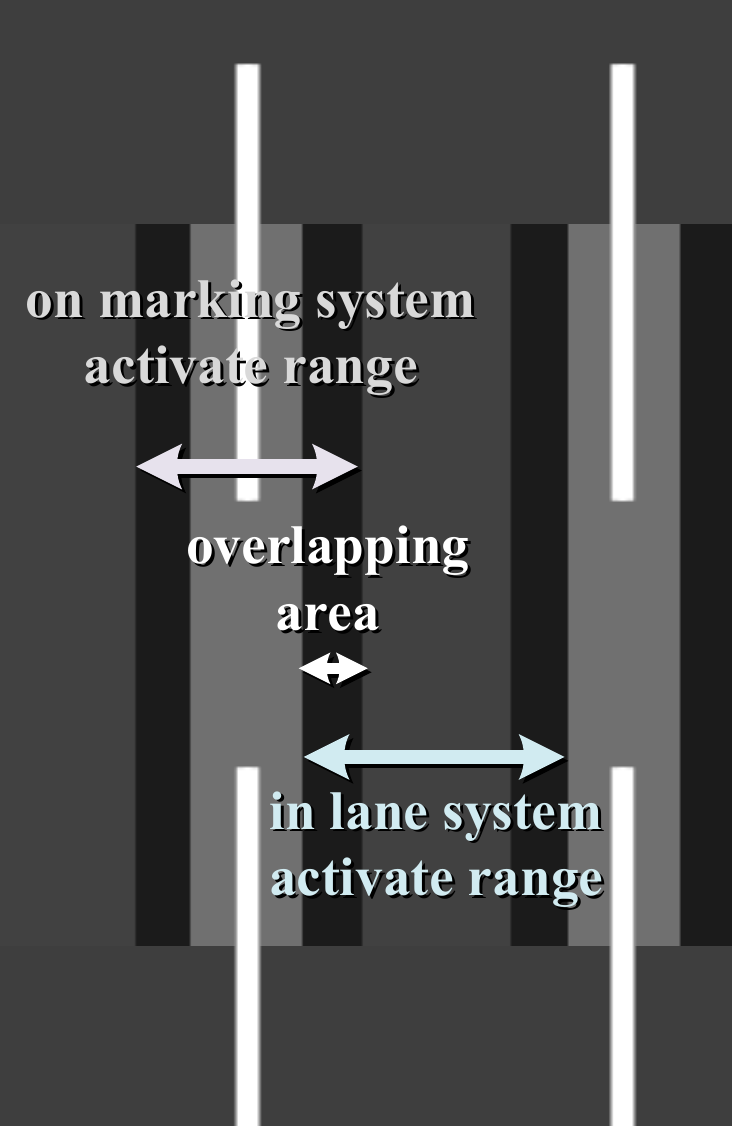}}

\vspace{-3mm}
\caption{{\bf Illustration of our affordance representation.}
A lane changing maneuver needs to traverse the ``in lane system" and the ``on marking system".
(f) shows the designated overlapping area used to enable smooth transitions.
}
\label{fig4}
\end{figure*}

\subsection{Related work}

Most autonomous driving systems from industry today are based on mediated perception approaches.
In computer vision, researchers have studied each task separately \cite{geiger2013vision}.
Car detection and lane detection are two key elements of an autonomous driving system.
Typical algorithms output bounding boxes on detected cars \cite{felzenszwalb2010object,lenz2011sparse}
and splines on detected lane markings \cite{aly2008real}.
However, these bounding boxes and splines are not the direct affordance information we use for driving.
Thus, a conversion is necessary which may result in extra noise.
Typical lane detection algorithms such as the one proposed in \cite{aly2008real} suffer from false detections.
Structures with rigid boundaries, such as highway guardrails or asphalt surface cracks, can be mis-recognized as lane markings.
Even with good lane detection results, critical information for car localization may be missing.
For instance, given that only two lane markings are usually detected reliably, it can be difficult to determine if a car is driving on the left lane or the right lane of a two-lane road.

To integrate different sources into a consistent world representation,
\cite{geiger20133d,zhang2013understanding}
proposed a probabilistic generative model that takes various detection results as inputs
and outputs the layout of the intersection and traffic details.

For behavior reflex approaches,
 \cite{pomerleau1989alvinn,pomerleau1992neural} are the seminal works that use a neural network to map images directly to steering angles.
More recently,
\cite{koutnik2013evolving} train a large recurrent neural network using a reinforcement learning approach. The network's function is the same as \cite{pomerleau1989alvinn,pomerleau1992neural}, mapping the image directly to the steering angles, with the objective to keep the car on track.
Similarly to us, they use the video game TORCS for training and testing.


In terms of deep learning for autonomous driving,
\cite{muller2005off} is a successful example of ConvNets-based behavior reflex approach. The authors propose an off-road driving robot DAVE that learns a mapping from images to a human driver's steering angles. After training, the robot demonstrates capability for obstacle avoidance.
\cite{hadsell2009learning} proposes an off-road driving robot with self-supervised learning ability for long-range vision.
In their system, a multi-layer convolutional network is used to classify an image segment as a traversable area or not.
For depth map estimation,
DeepFlow \cite{weinzaepfel2013deepflow} uses ConvNets to achieve very good results for driving scene images on the KITTI dataset \cite{geiger2013vision}.
For image features,
deep learning also demonstrates significant improvement \cite{krizhevsky2012imagenet,girshick2013rich,erhan2013scalable} over hand-crafted features, such as GIST \cite{oliva2001modeling}.
In our experiments, we will make a comparison between learned ConvNet features and GIST for direct perception in driving scenarios.

\section{Learning affordance for driving perception}


To efficiently implement and test our approach, we use the open source driving game TORCS (The Open Racing Car Simulator) \cite{TORCS}, which is widely used for AI research.
From the game engine, we can collect critical indicators for driving, e.g. speed of the host car,
the host car's relative position to the road's central line, the distance to the preceding cars. 
In the training phase, we manually drive a ``label collecting car" in the game to collect screenshots (first person driving view)
and the corresponding ground truth values of the selected affordance indicators.
This data is stored and used to train a model to estimate affordance in a supervised learning manner.
In the testing phase, at each time step, the trained model takes a driving scene image from the game and estimates the affordance indicators for driving. A driving controller processes the indicators and computes the steering and acceleration/brake commands.
The driving commands are then sent back to the game to drive the host car.
Ground truth labels are also collected during the testing phase to evaluate the system's performance.
In both the training and testing phase, traffic is configured by putting a number of pre-programmed AI cars on road.

\subsection{Mapping from an image to affordance}

We use a state-of-the-art deep learning ConvNet as
our direct perception model to map an image to the affordance indicators.
In this paper, we focus on highway driving with multiple lanes.
From an ego-centric point of view,
the host car only needs to concern the traffic in its current lane and the two adjacent (left/right) lanes when making decisions.
Therefore, we only need to model these three lanes.
We train a single ConvNet to handle three lane configurations together: a road of one lane, two lanes, or three lanes.
Shown in Figure~\ref{fig2} are the typical cases we are dealing with.
Occasionally the car has to drive on lane markings, and in such situations only the lanes on each side of the lane marking need to be monitored, as shown in Figure~\ref{fig3a} and \ref{fig3b}.

Highway driving actions can be categorized into two major types:
1) following the lane center line,
and 2) changing lanes or slowing down to avoid collisions with the preceding cars.
To support these actions,
we define our system to have two sets of representations
under two coordinate systems: ``in lane system" and ``on marking system".
To achieve two major functions, lane perception and car perception,
we propose three types of indicators to represent driving situations:
heading angle,
the distance to the nearby lane markings,
and the distance to the preceding cars.
In total, we propose 13 affordance indicators as our driving scene representation, illustrated in Figure~\ref{fig4}.
A complete list of the affordance indicators is enumerated in Figure~\ref{fig:list}.
They are the output of the ConvNet as our affordance estimation and the input of the driving controller.

\begin{figure}[t]
\begin{footnotesize}
\hrule
\vspace{1mm}
\textbf{always:}\\
\hspace*{1.5em}1) angle: angle between the car's heading and the tangent of the road\\
\textbf{``in lane system", when driving in the lane:}\\
\hspace*{1.5em}2) toMarking\_LL: distance to the left lane marking of the left lane\\
\hspace*{1.5em}3) toMarking\_ML: distance to the left lane marking of the current lane\\
\hspace*{1.5em}4) toMarking\_MR: distance to the right lane marking of the current lane\\
\hspace*{1.5em}5) toMarking\_RR: distance to the right lane marking of the right lane\\
\hspace*{1.5em}6) dist\_LL: distance to the preceding car in the left lane\\
\hspace*{1.5em}7) dist\_MM: distance to the preceding car in the current lane\\
\hspace*{1.5em}8) dist\_RR: distance to the preceding car in the right lane\\
\textbf{``on marking system", when driving on the lane marking:}\\
\hspace*{1.5em}9) toMarking\_L: distance to the left lane marking\\
\hspace*{1.5em}10) toMarking\_M: distance to the central lane marking\\
\hspace*{1.5em}11) toMarking\_R: distance to the right lane marking\\
\hspace*{1.5em}12) dist\_L: distance to the preceding car in the left lane\\
\hspace*{1.5em}13) dist\_R: distance to the preceding car in the right lane\\
\vspace{-2mm}\hrule
\end{footnotesize}

\vspace{2mm}
\caption{\bf Complete list of affordance indicators in our direct perception representation.}
\label{fig:list}
\vspace{-4mm}
\end{figure}

%

The ``in lane system" and ``on marking system" are activated under different conditions. To have a smooth transition, we define an overlapping area, where both systems are active. The layout is shown in Figure~\ref{fig5}.

Except for heading angle, all the indicators may output an inactive state.
There are two cases in which a indicator will be inactive:
1) when the car is driving in either the ``in lane system" or ``on marking system" and the other system is deactivated, then all the indicators belonging to that system are inactive.
2) when the car is driving on boundary lanes (left most or right most lane), and there is either no left lane or no right lane, then the indicators corresponding to the non-existing adjacent lane are inactive.
According to the indicators' value and active/inactive state, the host car can be accurately localized on the road.


\subsection{Mapping from affordance to action}

The steering control is computed using the car's position and pose, and the goal is to minimize the gap between the car's current position and the center line of the lane.
Defining $dist\_center$ as the distance to the center line of the lane, we have:
\begin{equation}
steerCmd=C*(angle-dist\_center/road\_width)
\end{equation}

\noindent
where $C$ is a coefficient that varies under different driving conditions, and $angle \in [-\pi,\pi]$. When the car changes lanes, the center line switches from the current lane to the objective lane.
The pseudocode describing the logic of the driving controller is listed in Figure~\ref{code}.


At each time step, the system computes $desired\_speed$. A controller makes the actual speed follow the $desired\_speed$ by controlling the acceleration/brake. The baseline $desired\_speed$ is 72 km/h. If the car is turning, a $desired\_speed$ drop is computed according to the past few steering angles. 
If there is a preceding car in close range and a slow down decision is made, the $desired\_speed$ is also determined by the distance to the preceding car. To achieve car-following behavior in such situations, we implement the optimal velocity car-following model \cite{newell1961nonlinear} as:
\begin{equation}
v(t)=v_{max}(1-\textrm{exp}(-\frac{c}{v_{max}}dist(t)-d))
\end{equation}
where $dist(t)$ is the distance to the preceding car, $v_{max}$ is the largest allowable speed, $c$ and $d$ are coefficients to be calibrated.
With this implementation, the host car can achieve stable and smooth car-following under a wide range of speeds and even make a full stop if necessary.

\begin{figure}[t]

\footnotesize
\hrule

\vspace{0.2cm}

\noindent
\textbf{while} (in autonomous driving mode) \\ 
\hspace*{1em} ConvNet outputs affordance indicators\\
\noindent
\hspace*{1em} check availability of both the left and right lanes\\
\noindent
\hspace*{1em} \textbf{if} (approaching the preceding car in the same lane)\\
\hspace*{2em}    \textbf{if} (left lane exists \textbf{and} available \textbf{and} lane changing allowable) \\
\hspace*{3em}        left lane changing decision made\\
\hspace*{2em}    \textbf{else if} (right lane exists \textbf{and} available \textbf{and} lane changing allowable) \\
\hspace*{3em}        right lane changing decision made\\
\hspace*{2em}    \textbf{else}\\
\hspace*{3em}        slow down decision made\\
\noindent
\hspace*{1em} \textbf{if} (normal driving)\\
\hspace*{2em}    $center\_line$= center line of current lane\\
\hspace*{1em} \textbf{else if} (left/right lane changing)\\
\hspace*{2em}    $center\_line$= center line of objective lane\\
\noindent
\hspace*{1em} compute steering command\\
\hspace*{1em} compute $desired\_speed$\\
\hspace*{1em} compute acceleration/brake command based on $desired\_speed$\\
\vspace{-2mm}\hrule

\vspace{2mm}
\caption{\bf Controller logic.}
\label{code}
\vspace{-4mm}
\end{figure}

\section{Implementation}

Our direct perception ConvNet is built upon Caffe \cite{jia2014caffe}, and we use the standard AlexNet architecture \cite{krizhevsky2012imagenet}. There are 5 convolutional layers followed by 4 fully connected layers with output dimensions of 4096, 4096, 256, and 13, respectively. Euclidian loss is used as the loss function. Because the 13 affordance indicators have various ranges, we normalize them to the range of $[0.1,0.9]$.

\begin{figure}[t]
  \centering
    \includegraphics[width=1\linewidth]{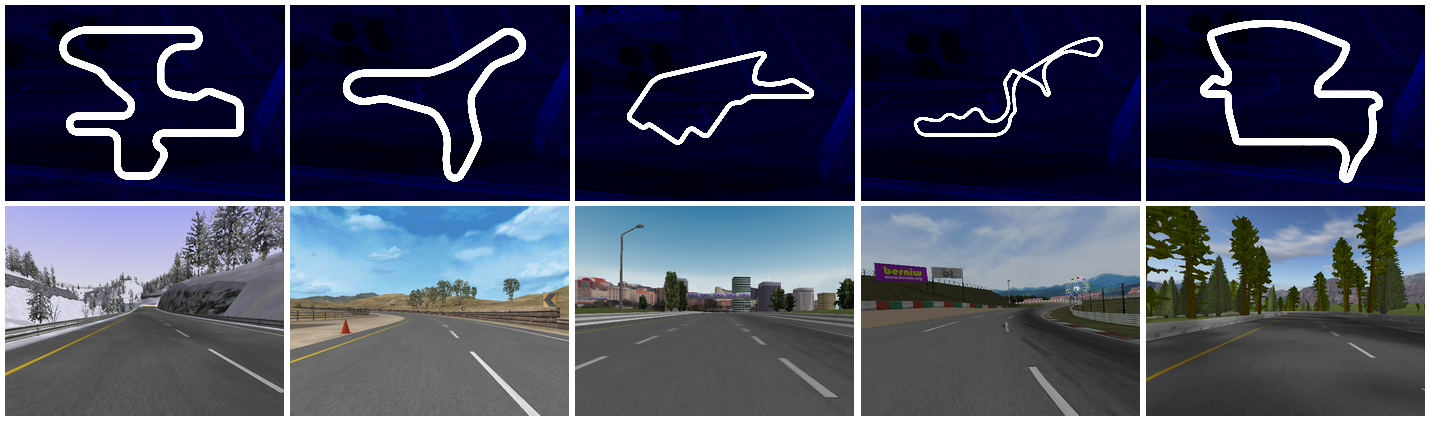}

    \vspace{-2mm}
    \caption{{\bf Examples of the 7 tracks used for training.} Each track is customized to the configuration of one-lane, two-lane, and three-lane with multiple asphalt darkness levels. The rest of the tracks are used in the testing set. }
    \label{fig6}

    \vspace{2mm}
    \includegraphics[width=1\linewidth]{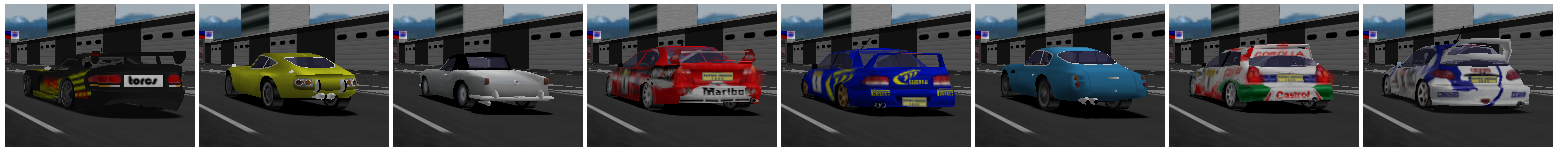}

    \vspace{-2mm}
    \caption{{\bf Examples of the 22 cars used in the training set}. The rest of the cars are used in the testing set. }
    \label{fig8}

\end{figure}

We select 7 tracks and 22 traffic cars in TORCS, shown in Figure~\ref{fig6} and Figure~\ref{fig8}, to generate the training set. We replace the original road surface textures in TORCS with over 30 customized asphalt textures of various lane configurations and asphalt darkness levels. 
We also program different driving behaviors for the traffic cars to create different traffic patterns.
We manually drive a car on each track multiple times to collect training data.
While driving, the screenshots are simultaneously down-sampled to $280\times210$ and stored in a database together with the ground truth labels. This data collection process can be easily automated by using an AI car. Yet, when driving manually, we can intentionally create extreme driving conditions (e.g. off the road, collide with other cars) to collect more effective training samples, which makes the ConvNet more powerful and significantly reduces the training time.

In total, we collect 484,815 images for training. The training procedure is similar to training an AlexNet on ImageNet data. The differences are: the input image has a resolution of $280\times210$ and is no longer a square image. We do not use any crops or a mirrored version. We train our model from scratch. We choose an initial learning rate of 0.01, and each mini-batch consists of 64 images randomly selected from the training samples. After 140,000 iterations, we stop the training process.

\begin{figure}[t]
  \centering
    \includegraphics[width=1\linewidth]{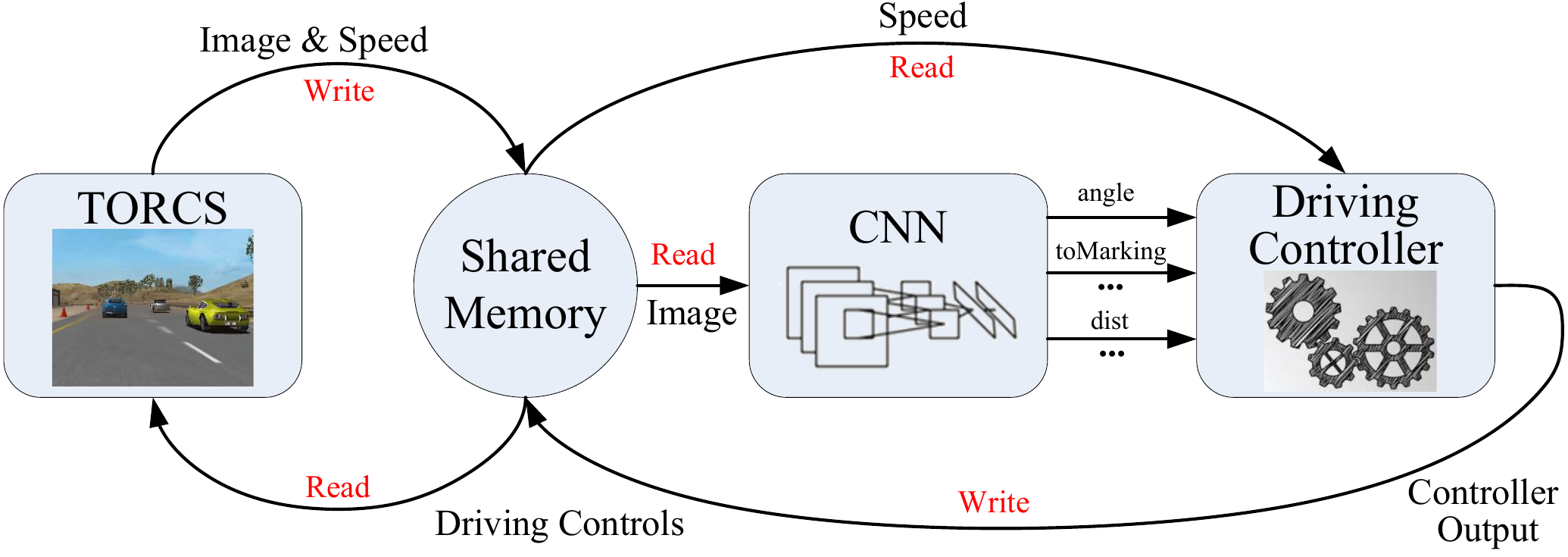}

    \vspace{-2mm}
    \caption{{\bf System architecture.} The ConvNet processes the TORCS image and estimates 13 indicators for driving. Based on the indicators and the current speed of the car, a controller computes the driving commands which will be sent back to TORCS to drive the host car in it. }
    \label{fig1}
\end{figure}

In the testing phase, when our system drives a car in TORCS, the only information it accesses is the front facing image and the speed of the car. Right after the host car overtakes a car in its left/right lane, it cannot judge whether it is safe to move to that lane, simply because the system cannot see things behind. To solve this problem, we make an assumption that the host car is faster than the traffic. Therefore if sufficient time has passed since its overtaking (indicated by a timer), it is safe to change to that lane. 
The control frequency in our system for TORCS is 10Hz, which is sufficient for driving below 80 km/h. A schematic of the system is shown in Figure~\ref{fig1}.

\section{TORCS evaluation}
We first evaluate our direct perception model on the TORCS driving game.
Within the game, the ConvNet output can be visualized and used by the controller to drive the host car. To measure the estimation accuracy of the affordance indicators, we construct a testing set consisting of tracks and cars not included in the training set.

In the aerial TORCS visualization (Figure~\ref{fig10}, right), we treat the host car as the reference object. As its vertical position is fixed, it moves horizontally with a heading computed from $angle$. Traffic cars only move vertically. We do not visualize the curvature of the road, so the road ahead is always represented as a straight line. Both the estimation (empty box) and the ground truth (solid box) are displayed.

\subsection{Qualitative assessment}
Our system can drive very well in TORCS without any collision. In some lane changing scenarios, the controller may slightly overshoot, but it quickly recovers to the desired position of the objective lane's center. As seen in the TORCS visualization, the lane perception module is pretty accurate, and the car perception module is reliable up to 30 meters away. In the range of 30 meters to 60 meters, the ConvNet output becomes noisier. In a $280\times210$ image, when the traffic car is over 30 meter away, it actually appears as a very tiny spot, which makes it very challenging for the network to estimate the distance. However, because the speed of the host car does not exceed 72 km/h in our tests, reliable car perception within 30 meters can guarantee satisfactory control quality in the game.

To maintain smooth driving, our system can tolerate moderate error in the indicator estimations. The car is a continuous system, and the controller is constantly correcting its position. Even with some scattered erroneous estimations, the car can still drive smoothly without any collisions.

\subsection{Comparison with baselines}
To quantitatively evaluate the performance of the TORCS-based direct perception ConvNet, we compare it with three baseline methods. We refer to our model as \textbf{``ConvNet full"} in the following comparisons.

\vspace{-3mm}\paragraph{1) Behavior reflex ConvNet:} The method directly maps an image to steering using a ConvNet. 
We train this model on the driving game TORCS using two settings: (1) The training samples (over 60,000 images) are all collected while driving on an empty track; the task is to follow the lane. (2) The training samples (over 80,000 images) are collected while driving in traffic; the task is to follow the lane, avoid collisions by switching lanes, and overtake slow preceding cars. 
The video in our project website shows the typical performance. 
For (1), the behavior reflex system can easily follow empty tracks. For (2), when testing on the same track where the training set is collected, the trained system demonstrates some capability at avoiding collisions by turning left or right. However, the trajectory is erratic. The behavior is far different from a normal human driver and is unpredictable - the host car collides with the preceding cars frequently.

\begin{figure}[t]
  \centering
    \includegraphics[width=0.7\linewidth]{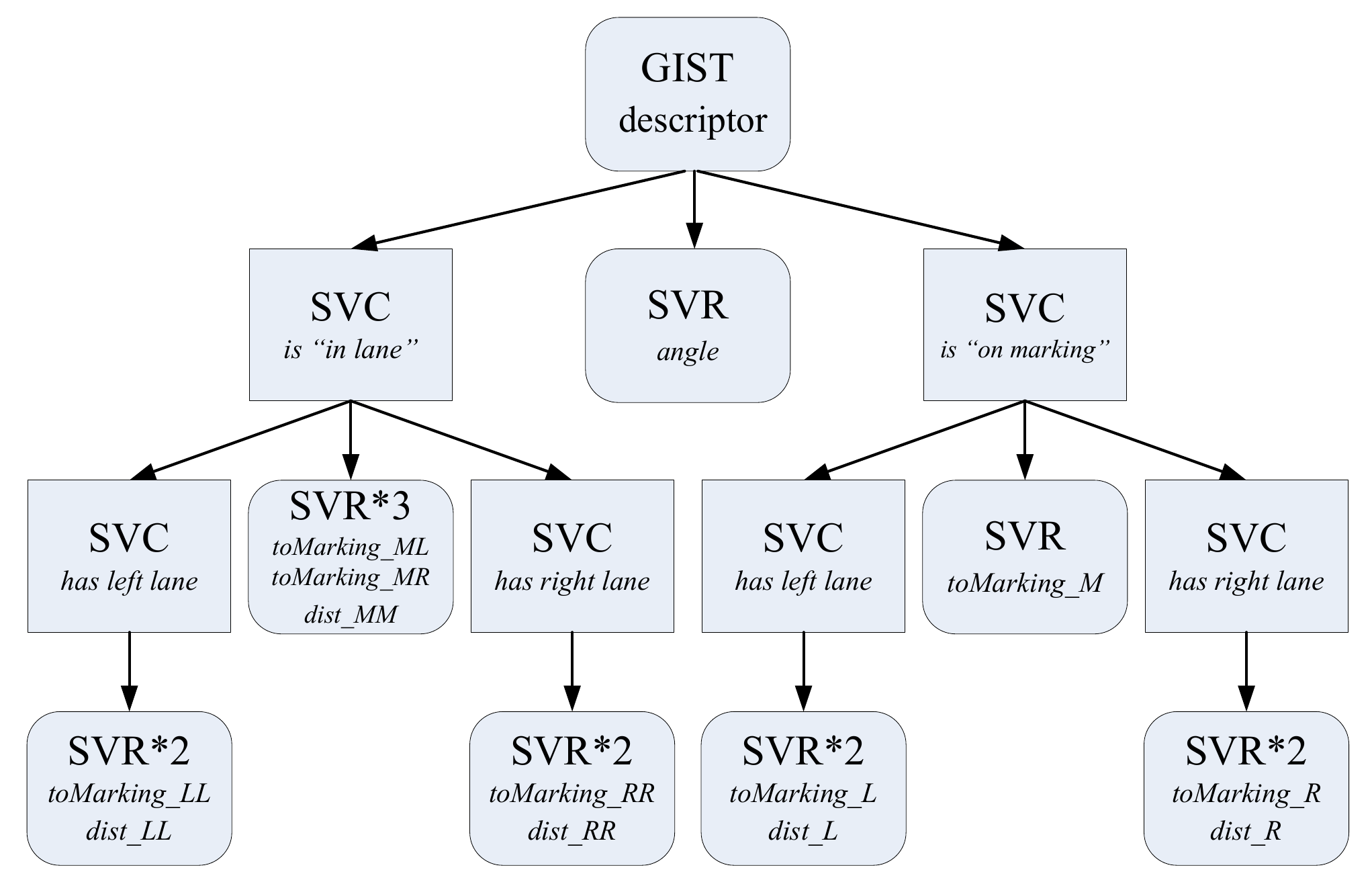}
    \caption{{\bf GIST baseline.} Procedure of mapping GIST descriptor to the 13 affordance indicators for driving using SVR and SVC.}
    \label{fig12}
\end{figure}

\vspace{-3mm}\paragraph{2) Mediated perception (lane detection):} We run the Caltech lane detector \cite{aly2008real} on TORCS images. Because only two lanes can be reliably detected, we map the coordinates of spline anchor points of the top two detected lane markings to the lane-based affordance indicators. We train a system composed of 8 Support Vector Regression (SVR) and 6 Support Vector Classification (SVC) models (using libsvm \cite{chang2011libsvm}) to implement the mapping (a necessary step for mediated perception approaches). The system layout is similar to the GIST-based system (next section) illustrated in Figure~\ref{fig12}, but without car perception.

Because the Caltech lane detector is a relatively weak baseline, to make the task simpler, we create a special training set and testing set. Both the training set (2430 samples) and testing set (2533 samples) are collected from the same track (not among the 7 training tracks for ConvNet) without traffic, and in a finer image resolution of $640\times480$. We discover that, even when trained and tested on the same track, the Caltech lane detector based system still performs worse than our model.
We define our error metric as Mean Absolute Error (MAE) between the affordance estimations and ground truth distances. A comparison of the errors for the two systems is shown in Table~\ref{tab3}.


\vspace{-3mm}\paragraph{3) Direct perception with GIST:} We compare the hand-crafted GIST descriptor with the deep features learned by the ConvNet's convolutional layers in our model. A set of 13 SVR and 6 SVC models are trained to convert the GIST feature to the 13 affordance indicators defined in our system. The procedure is illustrated in Figure~\ref{fig12}. The GIST descriptor partitions the image into $4\times4$ segments.
Because the ground area represented by the lower $2\times4$ segments may be more relevant to driving, we try two different settings in our experiments: (1) convert the whole GIST descriptor, and (2) convert the lower $2\times4$ segments of GIST descriptor. We refer to these two baselines as \textbf{``GIST whole"} and \textbf{``GIST half"} respectively.

Due to the constraints of libsvm, training with the full dataset of 484,815 samples is prohibitively expensive. We instead use a subset of the training set containing 86,564 samples for training. Samples in the sub training set are collected on two training tracks with two-lane configurations. To make a fair comparison, we train another ConvNet on the same sub training set for 80,000 iterations (referred to as \textbf{``ConvNet sub"}). The testing set is collected by manually driving a car on three different testing tracks with two-lane configurations and traffic. It has 8,639 samples.

\begin{figure}[t]
    \subfloat[Autonomous driving in TORCS]{\label{fig10}\includegraphics[width=0.5\linewidth]{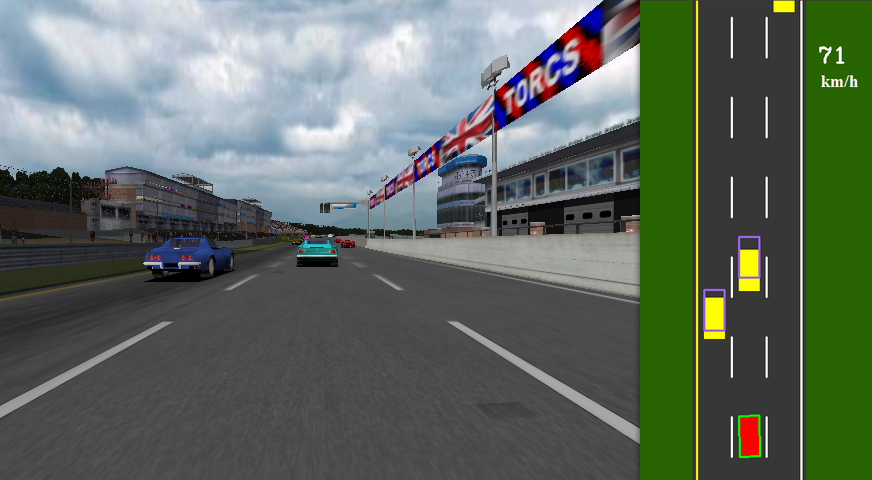}}~
    \subfloat[Testing on real video]{\label{fig11}\includegraphics[width=0.5\linewidth]{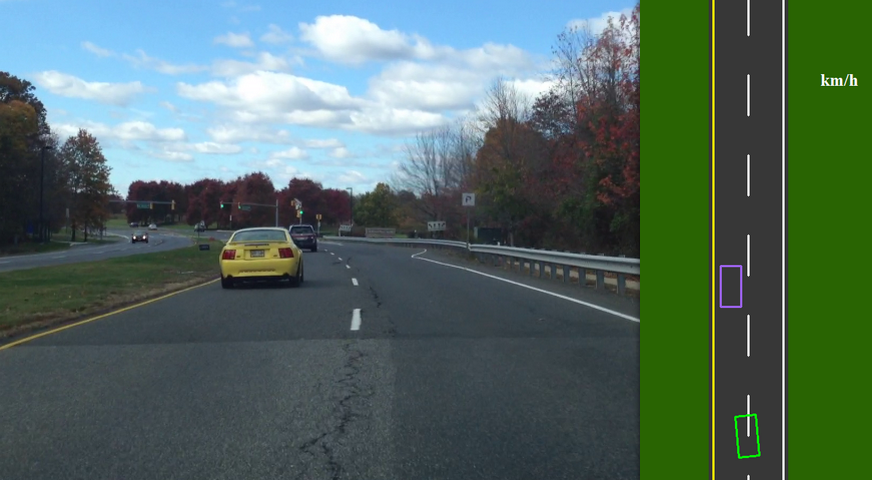}}

    \vspace{-3mm}
    \caption{{\bf Testing the TORCS-based system.} The estimation is shown as an empty box, while the ground truth is indicated by a solid box.
For testing on real videos, without the ground truth, we can only show the estimation.}
\end{figure}

\begin{table}[t]
\vspace{-2mm}
\centering
\footnotesize
\setlength{\tabcolsep}{2.5pt}
\begin{tabular}{l|c|c|c|c|c|c|c|c}
\hline
Parameter & angle & to\_LL & to\_ML & to\_MR & to\_RR & to\_L & to\_M & to\_R \\
\hline
Caltech lane & 0.048 & 1.673 & 1.179 & 1.084 & 1.220 & 1.113 & 1.060 & 0.895 \\
ConvNet full & 0.025 & 0.260 & 0.197 & 0.179 & 0.239 & 0.291 & 0.262 & 0.231 \\
\hline
\end{tabular}

\vspace{-2mm}
\caption{Mean Absolute Error (angle is in radians, the rest are in meters) on the testing set for the Caltech lane detector baseline.}
\label{tab3}
\vspace{-1mm}
\end{table}

The results are shown in Table~\ref{tab1}. The $dist$ (car distance) errors are computed when the ground truth cars lie within [2, 50] meters ahead. Below two meters, cars in the adjacent lanes are not visually present in the image.

\begin{table*}[t]
\begin{center}
\footnotesize
\begin{tabular}{l|c|c|c|c|c|c|c|c|c|c|c|c|c}
\hline
Parameter & angle & to\_LL & to\_ML & to\_MR & to\_RR & dist\_LL & dist\_MM & dist\_RR & to\_L & to\_M & to\_R & dist\_L & dist\_R \\
\hline
GIST whole & 0.051 & 1.033 & 0.596 & 0.598 & 1.140 & 18.561 & 13.081 & 20.542 & 1.201 & 1.310 & 1.462 & 30.164 & 30.138 \\
GIST half & 0.055 & 1.052 & 0.547 & 0.544 & 1.238 & 17.643 & 12.749 & 22.229 & 1.156 & 1.377 & 1.549 & 29.484 & 31.394 \\
ConvNet sub & 0.043 & 0.253 & 0.180 & 0.193 & 0.289 & 6.168 & 8.608 & 9.839 & 0.345 & 0.336 & 0.345 & 12.681 & 14.782 \\
ConvNet full & 0.033 & 0.188 & 0.155 & 0.159 & 0.183 & 5.085 & 4.738 & 7.983 & 0.316 & 0.308 & 0.294 & 8.784 & 10.740 \\
\hline
\end{tabular}
\end{center}

\vspace{-6mm}
\caption{Mean Absolute Error (angle is in radians, the rest are in meters) on the testing set for the GIST baseline.}
\label{tab1}
\end{table*}







Results in Table~\ref{tab1} show that the ConvNet-based system works considerably better than the GIST-based system. By comparing ``ConvNet sub" and ``ConvNet full", it is clear that more training data is very helpful for increasing the accuracy of the ConvNet-based direct perception system.


\section{Testing on real-world data}

\subsection{Smartphone video}
We test our TORCS-based direct perception ConvNet on real driving videos taken by a smartphone camera. Although trained and tested in two different domains, our system still demonstrates reasonably good performance. The lane perception module works particularly well. The algorithm is able to determine the correct lane configuration, localize the car in the correct lane, and recognize lane changing transitions. The car perception module is slightly noisier, probably because the computer graphics model of cars in TORCS look quite different from the real ones.
Please refer to the video on our project website for the result. 
A screenshot of the system running on real video is shown in Figure~\ref{fig11}.
Since we do not have ground truth measurements, only the estimations are visualized.



\subsection{Car distance estimation on the KITTI dataset}
To quantitatively analyze how the direct perception approach works on real images, we train a different ConvNet on the KITTI dataset \cite{geiger2013vision}. The task is estimating the distance to other cars ahead.

\begin{figure}[t]
\vspace{-1mm}
\centering

\subfloat[]{\label{fig14a}\includegraphics[width=0.265\linewidth]{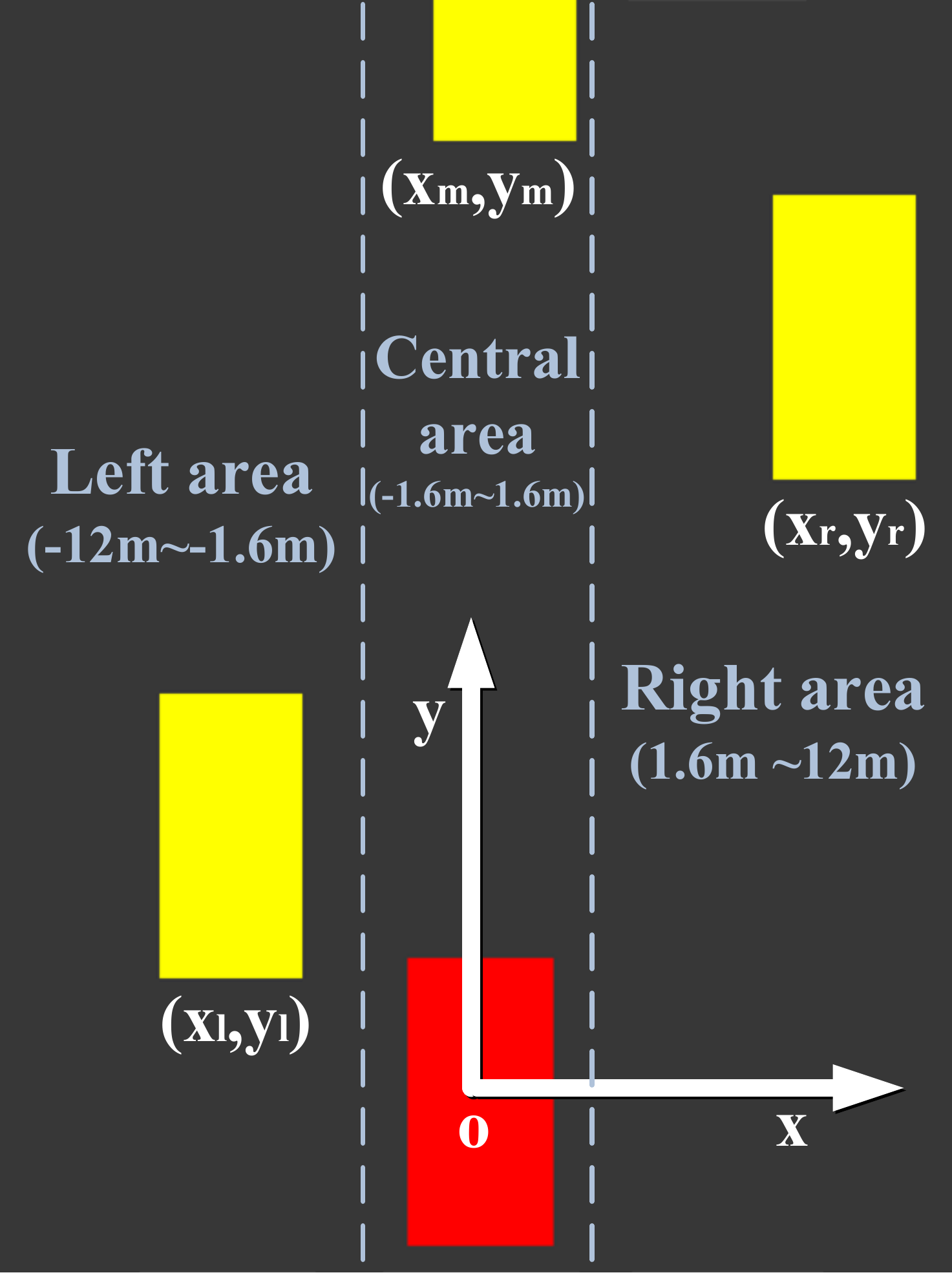}}~%
\subfloat[]{\label{fig14b}\includegraphics[width=0.745\linewidth]{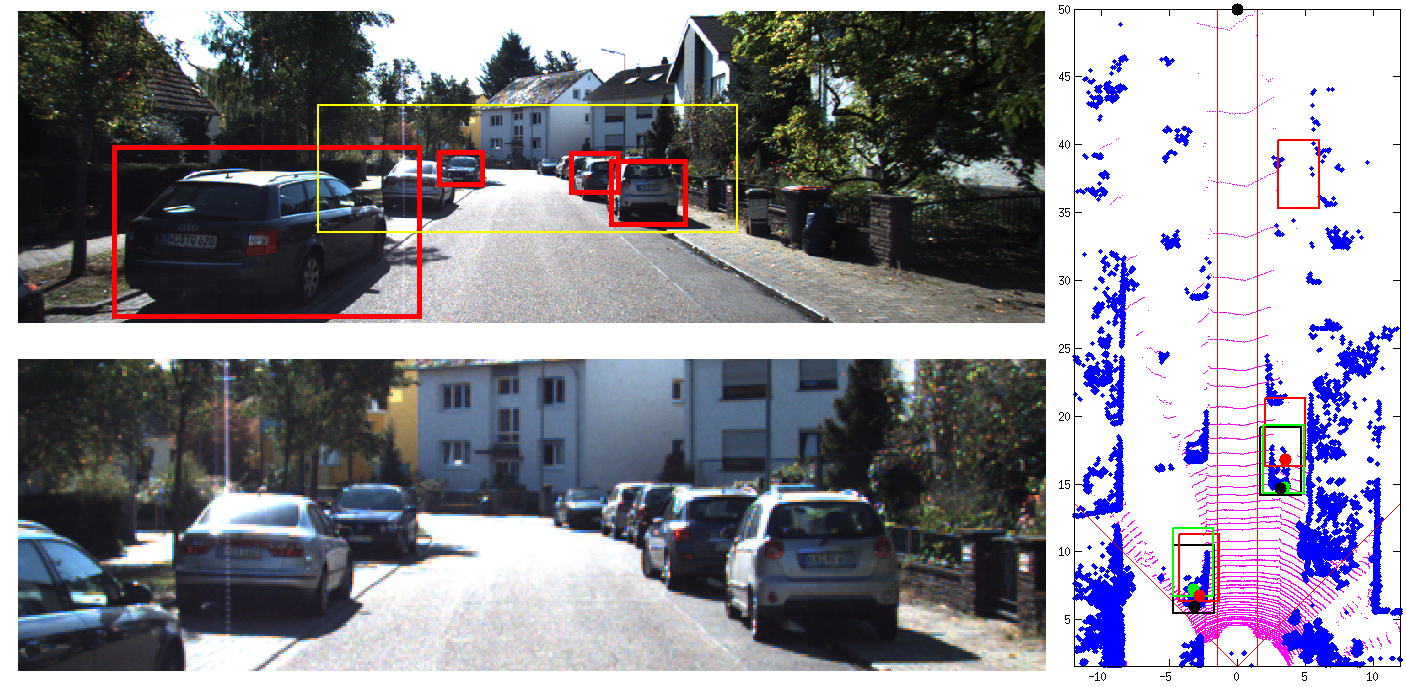}}

\vspace{-3mm}
\caption{{\bf Car distance estimation on the KITTI dataset.} (a) The coordinate system is defined relative to the host car. We partition the space into three areas, and the objective is to estimate the coordinate of the closest car in each area. (b) We compare our direct perception approach to the DPM-based mediated perception. The central crop of the KITTI image (indicated by the yellow box in the upper left image and shown in the lower left image) is sent to the far range ConvNet. The bounding boxes output by DPM are shown in red, as are its distance projections in the LiDAR visualization (right). The ConvNet outputs and the ground truth are represented by green and black boxes, respectively.}

\vspace{3.5mm}
\end{figure}

The KITTI dataset contains over 40,000 stereo image pairs taken by a car driving through European urban areas. Each stereo pair is accompanied by a Velodyne LiDAR 3D point cloud file. Around 12,000 stereo pairs come with official 3D labels for the positions of nearby cars, so we can easily extract the distance to other cars in the image. The settings for the KITTI-based ConvNet are altered from the previous TORCS-based ConvNet. In most KITTI images, there is no lane marking at all, so we cannot localize cars by the lane in which they are driving.
For each image, we define a 2D coordinate system on the zero height plane: the origin is the center of the host car, the $y$ axis is along the host car's heading, while the $x$ axis is pointing to the right of the host car (Figure~\ref{fig14a}). We ask the ConvNet to estimate the coordinate $(x,y)$ of the cars ``ahead" of the host car in this system.

There can be many cars in a typical KITTI image, but only those closest to the host car are critical for driving decisions. So it is not necessary to detect all the cars. We partition the space in front of the host car into three areas according to $x$ coordinate: 1) central area, $x \in [-1.6,1.6]$ meters, where cars are directly in front of the host car. 2) left area, $x \in [-12,1.6)$ meters, where cars are to the left of the host car. 3) right area, $x \in (1.6,12]$ meters, where cars are to the right of the host car. We are not concerned with cars outside this range. We train the ConvNet to estimate the coordinate $(x,y)$ of the closest car in each area (Figure~\ref{fig14a}). Thus, this ConvNet has 6 outputs.

Due to the low resolution of input images, cars far away can hardly be discovered by the ConvNet. We adopt a two-ConvNet structure. The close range ConvNet covers 2 to 25 meters (in the $y$ coordinate) ahead, and its input is the entire KITTI image resized to $497\times150$ resolution. The far range ConvNet covers 15 to 55 meters ahead, and its input is a cropped KITTI image covering the central $497\times150$ area. The final distance estimation is a combination of the two ConvNets' outputs. We build our training samples mostly from the KITTI officially labeled images, with some additional samples we labeled ourselves. The final number is around 14,000 stereo pairs. This is still insufficient to successfully train a ConvNet. We augment the dataset by using both the left camera and right camera images, mirroring all the images, and adding some negative samples that do not contain any car. Our final training set contains 61,894 images. Both ConvNets are trained on this set for 50,000 iterations. We label another 2,200 images as our testing set, on which we compute the numerical estimation error.

\subsection{Comparison with DPM-based baseline}

We compare the performance of our KITTI-based ConvNet with the state-of-the-art DPM car detector (a mediated perception approach). The DPM car detector is provided by \cite{geiger20133d} and is optimized for the KITTI dataset. We run the detector on the full resolution images and convert the bounding boxes to distance measurements by projecting the central point of the lower edge to the ground plane (zero height) using the calibrated camera model. The projection is very accurate given that the ground plane is flat, which holds for most KITTI images. DPM can detect multiple cars in the image, and we select the closest ones (one on the host car's left, one on its right, and one directly in front of it) to compute the estimation error. Since the images are taken while the host car is driving, many images contain closest cars that only partially appear in the left lower corner or right lower corner. DPM cannot detect these partial cars, while the ConvNet can better handle such situations. To make the comparison fair, we only count errors when the closest cars fully appear in the image. The error is computed when the traffic cars show up within 50 meters ahead (in the $y$ coordinate). When there is no car present, the ground truth is set as 50 meters. Thus, if either model has a false positive, it will be penalized. The Mean Absolute Error (MAE) for the $y$ and $x$ coordinate, and the Euclidian distance $d$ between the estimation and the ground truth of the car position are shown in Table~\ref{tab4}. A screenshot of the system is shown in Figure~\ref{fig14b}.

\begin{figure}[t]
  \centering
    \includegraphics[width=0.99\linewidth]{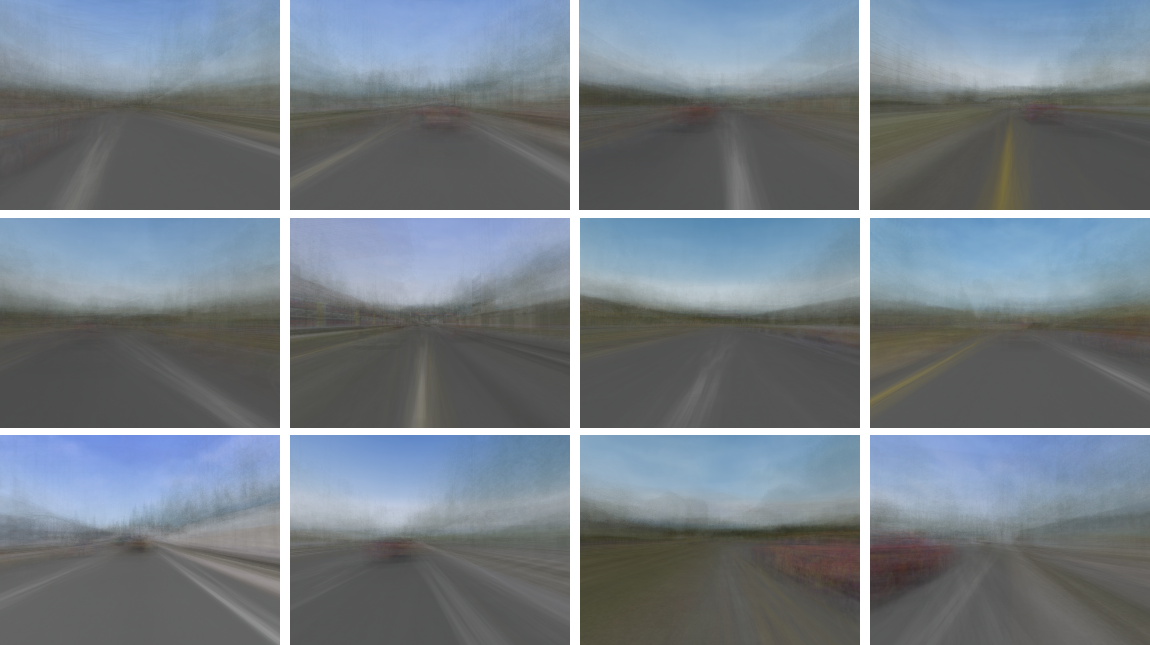}

    \vspace{-2mm}
    \caption{{\bf Activation patterns of neurons.} The neurons' activation patterns display strong correlations with the host car's heading, the location of lane markings, and traffic cars.}
    \label{fig15}
\vspace{-1mm}    
\end{figure}

\begin{table}[t]
\centering
\footnotesize
\setlength{\tabcolsep}{5pt}
\begin{tabular}{l|c|c|c||p{0.8cm}|p{0.8cm}|p{0.8cm}}
\hline
Parameter & $y$ & $x$ & $d$ & $y \setminus FP$ & $x \setminus FP$ & $d \setminus FP$ \\
\hline
ConvNet & 5.832 & 1.565 & 6.299 & 4.332 & 1.097 & 4.669 \\
DPM + Proj. & 5.824 & 1.502 & 6.271 & 5.000 & 1.214 & 5.331 \\
\hline
\end{tabular}

\vspace{-3mm}
\caption{Mean Absolute Error (in meters) on the KITTI testing set. Errors are computed by both penalizing (column 1$\sim$3) and not penalizing false positives (column 4$\sim$6).}
\label{tab4}

\end{table}

From Table~\ref{tab4}, we observe that our direct perception ConvNet has similar performance to the state-of-the-art mediated perception baseline. Due to the cluttered driving scene of the KITTI dataset, and the limited number of training samples, our ConvNet has slightly more false positives than the DPM baseline on some testing samples. If we do not penalize false positives, the ConvNet has much lower error than the DPM baseline, which means its direct distance estimations of true cars are more accurate than the DPM-based approach. From our experience, the false positive problem can be reduced by simply including more training samples. Note that the DPM baseline requires a flat ground plane for projection. If the host car is driving on some uneven road (e.g. hills), the projection will introduce a considerable amount of error. We also try building SVR regression models mapping the DPM bounding box output to the distance measurements.
But the regressors turn out to be far less accurate than the projection. 


\section{Visualization}

To understand how the ConvNet neurons respond to the input images, we can visualize the activation patterns. On an image dataset of 21,100 samples, for each of the 4,096 neurons in the first fully connected layer, we pick the top 100 images from the dataset that activate the neuron the most and average them to get an activation pattern for this neuron. In this way, we gain an idea of what this neuron learned from training. Figure \ref{fig15} shows several randomly selected averaged images. We observe that the neurons' activation patterns have strong correlation with the host car's heading, the location of the lane markings and the traffic cars. Thus we believe the ConvNet has developed task-specific features for driving.

\begin{figure}[t]
  \centering
    \includegraphics[width=1\linewidth]{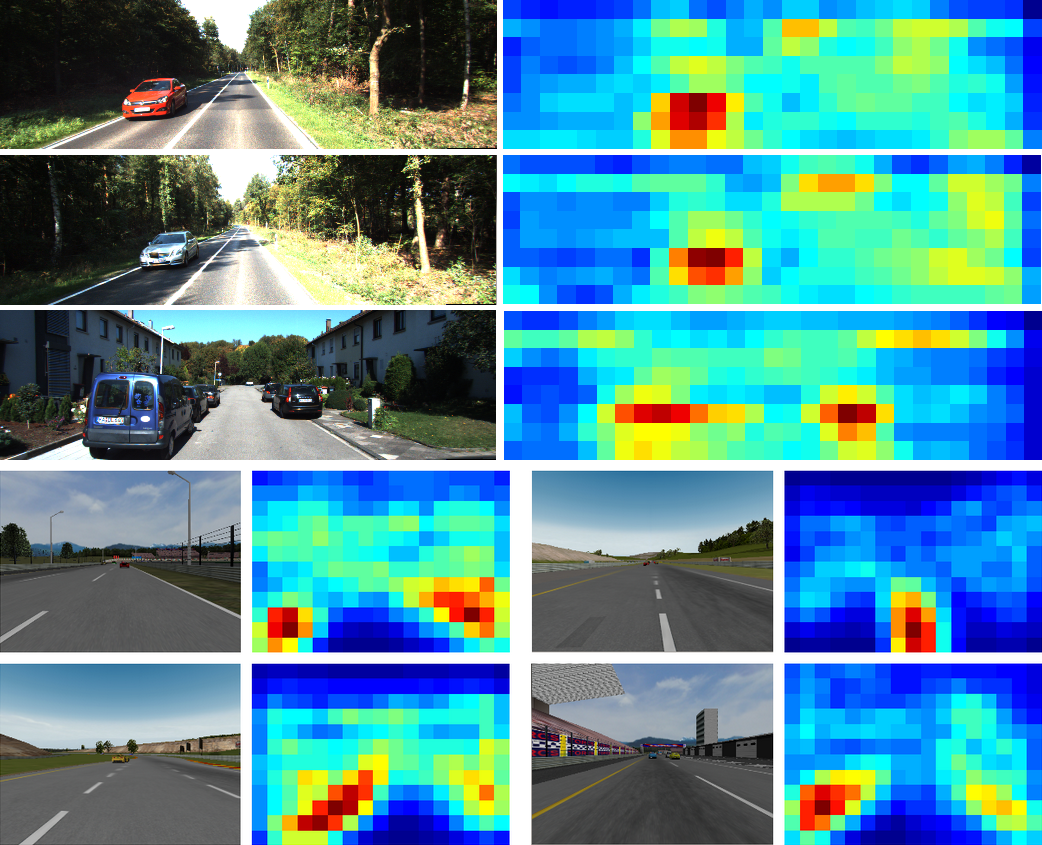}

    \vspace{-2mm}
    \caption{{\bf Response map} of our KITTI-based (Row 1-3) and TORCS-based (Row 4-5) ConvNets. The ConvNets have strong responses over nearby cars and lane markings. }
    \label{fig16}
\end{figure}

For a particular convolutional layer of the ConvNet, a response map can be generated by displaying the highest value among all the filter responses at each pixel. Because location information of objects in the original input image is preserved in the response map, we can learn where the salient regions of the image are for the ConvNet when making estimations for the affordance indicators. We show the response maps of the 4th convolutional layer of the close range ConvNet on a sample of KITTI testing images in Figure \ref{fig16}. We observe that the ConvNet has strong responses over the locations of nearby cars, which indicates that it learns to ``look" at these cars when estimating the distances. We also show some response maps of our TORCS-based ConvNet in the same figure. This ConvNet has very strong responses over the locations of lane markings.

\section{Conclusions}

In this paper, we propose a novel autonomous driving paradigm based on direct perception.
Our representation leverages a deep ConvNet architecture to estimate the affordance for driving actions
instead of parsing entire scenes (mediated perception approaches),
or blindly mapping an image directly to driving commands (behavior reflex approaches).
Experiments show that
our approach can perform well 
in both virtual and real environments.

\vspace{-3mm}
\paragraph{Acknowledgment.}
This work is partially supported by gift funds from Google, Intel Corporation and Project X grant to the Princeton Vision Group,
and a hardware donation from NVIDIA Corporation.

{\small
\bibliographystyle{ieee}
\bibliography{refs}
}

\end{document}